\documentclass[runningheads]{llncs}
\usepackage[T1]{fontenc}
\usepackage{graphicx}
\usepackage{verbatim}
\usepackage{booktabs}
\usepackage{xcolor} % for \textcolor
\usepackage{esvect} % for equations

\begin{document}
\title{Validation of Various Normalization Methods for Brain Tumor Segmentation: Can Federated Learning Overcome This Heterogeneity?}
\titlerunning{Validation of Various Normalization Methods for Brain Tumor Segmentation}

%
\begin{comment}  %% Removed for anonymized MICCAI 2025 submission
\author{First Author\inst{1}\orcidID{0000-1111-2222-3333} \and
Second Author\inst{2,3}\orcidID{1111-2222-3333-4444} \and
Third Author\inst{3}\orcidID{2222--3333-4444-5555}}
%
\authorrunning{F. Author et al.}
% First names are abbreviated in the running head.
% If there are more than two authors, 'et al.' is used.
%
\institute{Princeton University, Princeton NJ 08544, USA \and
Springer Heidelberg, Tiergartenstr. 17, 69121 Heidelberg, Germany
\email{lncs@springer.com}\\
\url{http://www.springer.com/gp/computer-science/lncs} \and
ABC Institute, Rupert-Karls-University Heidelberg, Heidelberg, Germany\\
\email{\{abc,lncs\}@uni-heidelberg.de}}

\end{comment}

\author{Jan Fiszer\inst{1,2}, Dominika Ciupek\inst{1}, Maciej Malawski\inst{1,2}}  %% Added for anonymized MICCAI 2025 submission
\authorrunning{Fiszer, Ciupek, Malawski et al.}
\institute{1. Sano Centre for Computational Medicine, Krakow, Poland \email{j.fiszer@sanoscience.org}, \email{d.ciupek@sanoscience.org} \\ 2. AGH University of Krakow, Poland}

\maketitle              % typeset the header of the contribution
\begin{abstract}
Deep learning (DL) has been increasingly applied across various fields, including medical imaging, where it often outperforms traditional approaches. However, DL requires large amounts of data, which raises many challenges related to data privacy, storage, and transfer. Federated learning (FL) is a training paradigm that overcomes these issues, though its effectiveness may be reduced when dealing with non-independent and identically distributed (non-IID) data. This study simulates non-IID conditions by applying different MRI intensity normalization techniques to separate data subsets, reflecting a common cause of heterogeneity. These subsets are then used for training and testing models for brain tumor segmentation. The findings provide insights into the influence of the MRI intensity normalization methods on segmentation models, both training and inference. Notably, the FL methods demonstrated resilience to inconsistently normalized data across clients, achieving the 3D Dice score of 92\%, which is comparable to a centralized model (trained using all data). These results indicate that FL is a solution to effectively train high-performing models without violating data privacy, a crucial concern in medical applications.

% The abstract should briefly summarize the contents of the paper in 150--250 words.  If you are to include a link to your Repository, please make sure it is anonymized for the double-blind review phase.

% Authors must provide keywords and are not allowed to remove this Keyword section.

\end{abstract}

\section{Introduction}
Artificial intelligence is transforming healthcare by improving diagnostic accuracy, streamlining workflows, and aiding clinical decisions \cite{alowais2023revolutionizing}. 
% \cite{for more https://chatgpt.com/c/684fe61a-7c2c-800d-a785-cad22d62b481}.  
In particular, deep learning excels at finding complex patterns from large datasets, achieving top performance in tasks such as tumor detection or organ segmentation \cite{fu2021review,liu2023deep}. However, its clinical adoption is limited by the need for vast amounts of high-quality annotated data—challenging to obtain due to strict privacy regulations and restricted data sharing.
% Artificial intelligence has become a transformative force in healthcare, particularly in medical imaging, where it has the potential to enhance diagnostic accuracy, streamline workflows, and support clinical decision-making. Among the many AI techniques, deep learning (DL) has stood out for its ability to learn complex patterns from large datasets, yielding state-of-the-art performance in tasks such as tumor detection and organ segmentation. However, the widespread adoption of DL in clinical practice is constrained by a critical dependency: access to large volumes of high-quality annotated data. In healthcare, where patient privacy is paramount and data sharing is heavily restricted, this poses a significant barrier.

To address these challenges, federated learning (FL)\cite{mcmahan2017originfl} has emerged as a promising paradigm. Rather than requiring data to be centralized, FL enables institutions to collaboratively train a shared model while keeping patient data local. This approach seeks to leverage large-scale data while maintaining privacy. However, FL introduces new complexities, especially in scenarios where data across institutions (\textit{FL clients}) is not uniformly distributed \cite{zhu2021flnoniddsurvey}. Differences in imaging protocols, scanner hardware, and preprocessing techniques can lead to variations that degrade model performance.
Additionally, in magnetic resonance imaging (MRI), various intensity normalization\footnote{In the following, the word \textit{normalization} refers to the intensity normalization.} techniques have been developed \cite{nyul1999standardizing,reinhold2019sythesis_normaliation,shinohara2014whitestripe}, but none of them have been established as the superior one, becoming an additional source of heterogeneity.

This article delves into one such real-world complexity: the impact of heterogeneous MRI intensity normalization techniques on performance in brain tumor segmentation. By simulating client-specific preprocessing, the study explores how inconsistencies in data normalization can affect model training and reliability. The study evaluated different MRI normalization methods across separate data subsets, assessing their influence on the accuracy of brain tumor segmentation models. The results verify the proficiency of five normalization methods for the training and inference of segmentation models. Further, it presents the capabilities of FL models to achieve near-parity with centralized models, even under non-independent and identically distributed (non-IID) conditions. 
% \textcolor{lightgray}{These findings contribute to the growing body of evidence supporting FL as a viable path forward for secure, high-performance medical AI.}

The previous work investigated the influence of different normalization methods \cite{cerebellum_segmenetiatoionjacobsen2019,reinhold2019sythesis_normaliation} on deep learning tasks, but to the best of our knowledge, it hasn't been validated for brain tumor segmentation. Further, there is a wide variety of experiments with FL for heterogeneous and non-IID data, but no publications were found regarding exactly the behavior of FL with various normalization methods between clients, a particular scenario of attribute-skew\cite{zhu2021flnoniddsurvey}.  

The main contributions are:
\begin{enumerate}
    \item Verification of five normalization methods for MRI from the automatic brain tumor segmentation,
    \item Proof of robustness of FL methods against variously normalized models.
\end{enumerate}
\section{Materials \& methods}
\begin{figure}
\includegraphics[width=\textwidth]{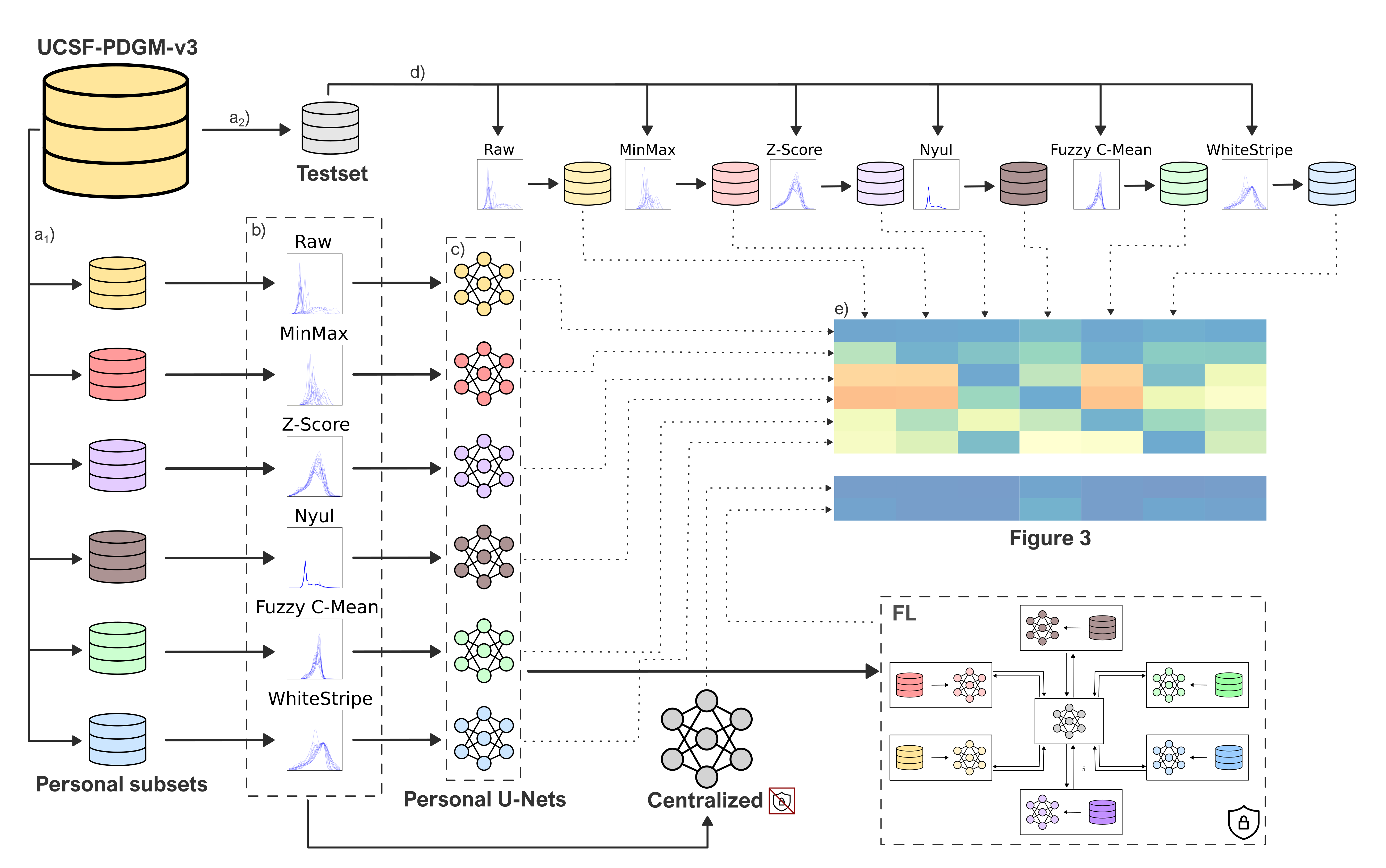}
\caption{The \textit{big picture} of the pipeline used, explaining the structure of the results table in Figure \ref{fig:dice-table}. Visualization of the main steps and their dependencies, including: a) the UCSF-PDGM-v3 dataset split, b) subsets normalization, c) models training, d) test sets normalization, and e) resulting evaluation table.} \label{fig:pipeline}
\end{figure}
\subsection{Data}
\subsubsection{Dataset}

% \textbf{Coppied from FL-translation}

% \textit{Few general sentences about the dataset}
% \textcolor{gray}{(UCSF-PDGM-v3) (Calabrese et al., 2022): 150 subjects (63F/87M), aged 57±15, with histopathologically confirmed gliomas according to WHO classification (grades 2-4). All subjects were scanned using a 3T Discovery 750 GE scanner (GE, Waukesha, WI) with a dedicated eight-channel head coil. The dataset includes T1-/T2-weighted/FLAIR scans acquired using 3D IR-SPGR/3D FSE/3D FSE protocols, respectively. Other acquisition parameters were: voxel resolution: 1  1  1 mm3 for T1-weighted scans and 1.2  1.2  1.2 mm3 for T2-weighted/FLAIR.}
The dataset used in this study was UCSF-PDGM-v3 \cite{calabrese2022ucsf-ds}, which includes 495 subjects diagnosed with WHO grade 2 to 4 gliomas confirmed by histopathology. Each subject had multiple MRI modalities acquired using a 3T Discovery 750 GE scanner (GE, Waukesha, WI), along with expert-validated brain tumor segmentation masks. Only T1-weighted, T2-weighted, and FLAIR images were selected, as these are compatible with the normalization techniques being evaluated.
During manual filtering, two subjects were removed due to strong artifacts,  leaving 493 subjects for further processing.

The dataset was divided into six equally sized personal subsets (one for each normalization and one raw — not-normalized), with a total of 82 patients each. Afterward, each of the subsets was normalized by one of the normalization methods described in subsection \textit{Normalizations}. That procedure artificially generated six non-IID subsets, one for each client. That mimics a real-case scenario such that, during the federated learning process, collaborating institutions have very similar data, but different preprocessing methods (here, differently normalized). Furthermore, the subsets were divided into train, test, and validation sets in proportions 75:20:5, which consequently resulted in 61, 17, and 4 brain volumes for each of the sets (see Figure \ref{fig:pipeline}). Since the validation set was not particularly important, this amount (4 volumes) was sufficient. It was not relevant for federated learning and was just used to monitor the process of classical training. The test sets were used during federated learning for on-site client testing.

\begin{figure}
\includegraphics[width=\textwidth]{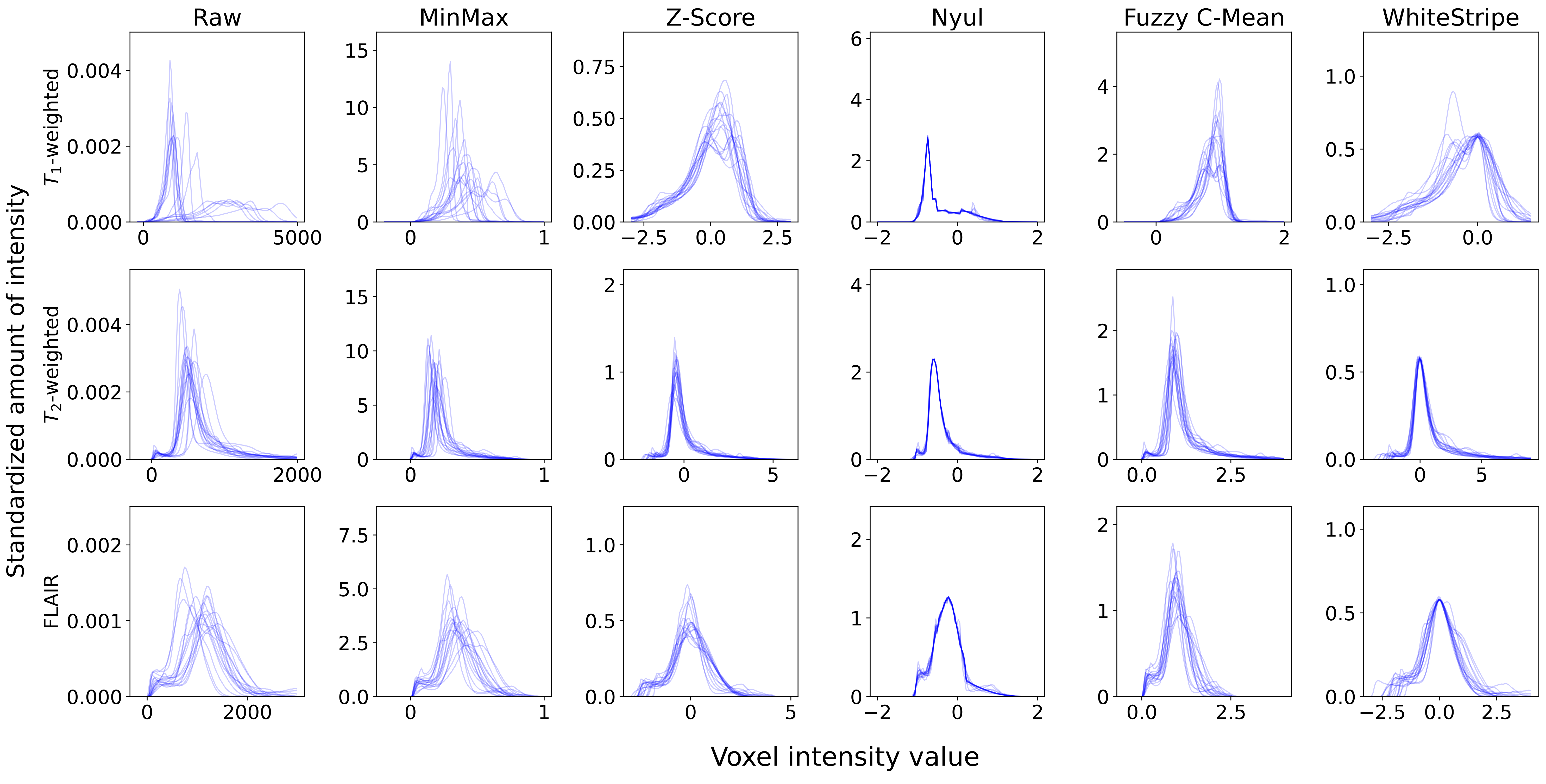}
\caption{Visualization of the histograms for all modalities for each of the normalization methods. Each line corresponds to the brain (without the background) voxels' intensity distribution. The 17 subjects from the common test set volumes were used for this visualization. } \label{fig_norm_histograms}
\label{fig:normalization}
\end{figure}

For proper evaluation, the common test sets originated from the same subset of data (same subjects), but were differently normalized (presented in Figure \ref{fig:normalization}). The common test sets also consisted of 17 patients, equal to 1466 brain slices (division illustrated in Figure \ref{fig:pipeline}) of which 1036 included brain tumors.

All volumes used were divided into 2D brain slices, resulting in 240$\times$240 images, as the segmentation model was working on 2D data (see Section \ref{sec:model-archi}). The slices were filtered to contain at least 20\% of brain pixels or for these having a brain tumor mask of at least 5\% of the brain pixels. The segmentation task was simplified by binarization of the target tumor mask, so instead of having three possible tumor compartments, there was just one.

\subsubsection{Normalizations}
\label{sec:normalization}
The artificial heterogeneity of the data was introduced by the use of varying normalization techniques for MR images: MinMax scaling, Z-score \cite{reinhold2019sythesis_normaliation}, Nyul \cite{nyul1999standardizing}, Fuzzy C-Mean \cite{reinhold2019sythesis_normaliation}, WhiteStripe \cite{shinohara2014whitestripe}. They were applied with the help of the Python package \texttt{intensity-normalization} \cite{reinhold2019sythesis_normaliation}. The Nyul method is Piecewise Linear Histogram Matching that learns the standard histogram and adjusts each volume to fit it. Fuzzy C-Mean (FCM) calculates the mean value of a specified tissue and uses it as a normalization factor. WhiteStripe performs Z-score normalization based on the white matter intensities. The hyperparameters used are the same as in the original publications \cite{nyul1999standardizing,shinohara2014whitestripe}.

% \textbf{Description of more complex methods}

\subsection{Segmentation model}
\subsubsection{Model architecture}
\label{sec:model-archi}
The utilized segmentation model is 2D U-Net \cite{ronneberger2015unet}, which, as input, takes 3 images (T1-weighted, T2-weighted, FLAIR) and outputs the brain tumor probability map. 
The encoder begins with a block containing 64 channels, progressively doubling the number of channels at each stage until it reaches 1024 at the bottleneck. The decoder then reverses this pattern, starting from the bottleneck and outputs the tumor mask. Unlike the classical implementation, here U-Net uses Group Normalization \cite{wu2018groupnorm} instead of Batch Normalization \cite{ioffe2015batchnorm}, since the latter harms federated learning \cite{wang2023bnharmsfl} and yielded worse performance. Additionally, a dropout \cite{srivastava2014dropout} layer was incorporated into the downsampling blocks with the dropout ratio of 0.3 (best performance among \{0.1, 0.3, 0.5\}).

\subsubsection{Training}
The neural network was trained using Adam optimizer \cite{kingma2014adam} with the learning rate value of 0.001 (best performing among \{0.01, 0.001, 0.0001\}) with \textit{Generalized Dice Loss} ($GDL$) function \cite{sudre2017generaliseddice}:
\begin{equation}
    GDL(P, T) = 1 - GDS(P, T)
\end{equation}
\begin{equation}
    GDS(P, T) = 2 \frac{\sum_{l=1}^{2} w_l \sum_n t_{ln} p_{ln}}{\sum_{l=1}^{2} w_l \sum_n (t_{ln} + p_{ln})}
    \label{eq:gen_dice}
\end{equation}
where $T$ is the target mask image (ground truth) and $P$ predicted probability values of the tumor throughout the image, with voxel/pixel values  $t_{ln}$ and $p_{ln}$, respectively. The weight $w_l$ balances the disproportion of the pixel quantity and is equal to $\frac{1}{\left(\sum_{n=1}^{N} t_{ln}\right)^2}$.  

This loss function handled the issue of a relatively small number of target pixels (sometimes even no target mask) by weighting inversely proportional to the number of brain pixels. For training, the soft Dice variant was used, meaning the values were not binarized in the calculation process.

For all classically (non-FL) trained models, the training was for 16 epochs, which led to sufficient convergence. Moreover, for the evaluation, the model with the lowest loss across the whole training process was taken, reducing the probability of overfitting. Any overfitting would be particularly undesirable due to the significant distribution skew between different test sets.

\subsubsection{Federated learning}
Regarding FL hyperparameters, the models were trained with 2 local epochs and 32 global rounds, and in every round, all clients were used. This setup led to stabilized convergence, and the low number of local epochs prevented local overfitting during rounds.
The evaluated aggregation methods were FedAvg and FedBN \cite{li2021fedbn}. FedAvg is the FL baseline, which calculates the weighted average based on the number of samples. However, because of the similar number of data samples for each client, the averaging weights were almost alike. FedBN aggregates the model similarly, but omits the normalization layers parameters, keeping them personalized for each client. Therefore, during the evaluation, personalized models were used for the corresponding dataset.
% \textbf{Describe FedBN and FedAvg}

\subsubsection{Quantitative evaluations}
For evaluation, the $GDS$ described by Equation \ref{eq:gen_dice} was used, but computed with inputs ($P$ and $T$) of the entire volumes (3D Dice variant). The network was deployed for all the subject slices, then the $GDS$ was calculated on the concatenated slices. Since the brain MRIs are always considered three-dimensional, that evaluation approach was more appropriate and comparable with different approaches. Furthermore, it eliminated the problem of low scores appearing for slices with only a few tumor pixels, which is particularly challenging for the network, missing spatial context (as a 2D U-Net). 

% Dice score, the standard binary segmentation evaluation metric, was used during model testings. Slightly different from the one in the loss function: a more standard version for comparison purposes concerning other publications. 
% \begin{equation}
%     dice = ...
% \end{equation}
% Here, a variant with binarization of the prediction was implemented. In addition, the smoothing dealt with cases without target brain tumor masks. 

\section{Results}
\begin{figure}[tb]
\includegraphics[width=\textwidth]{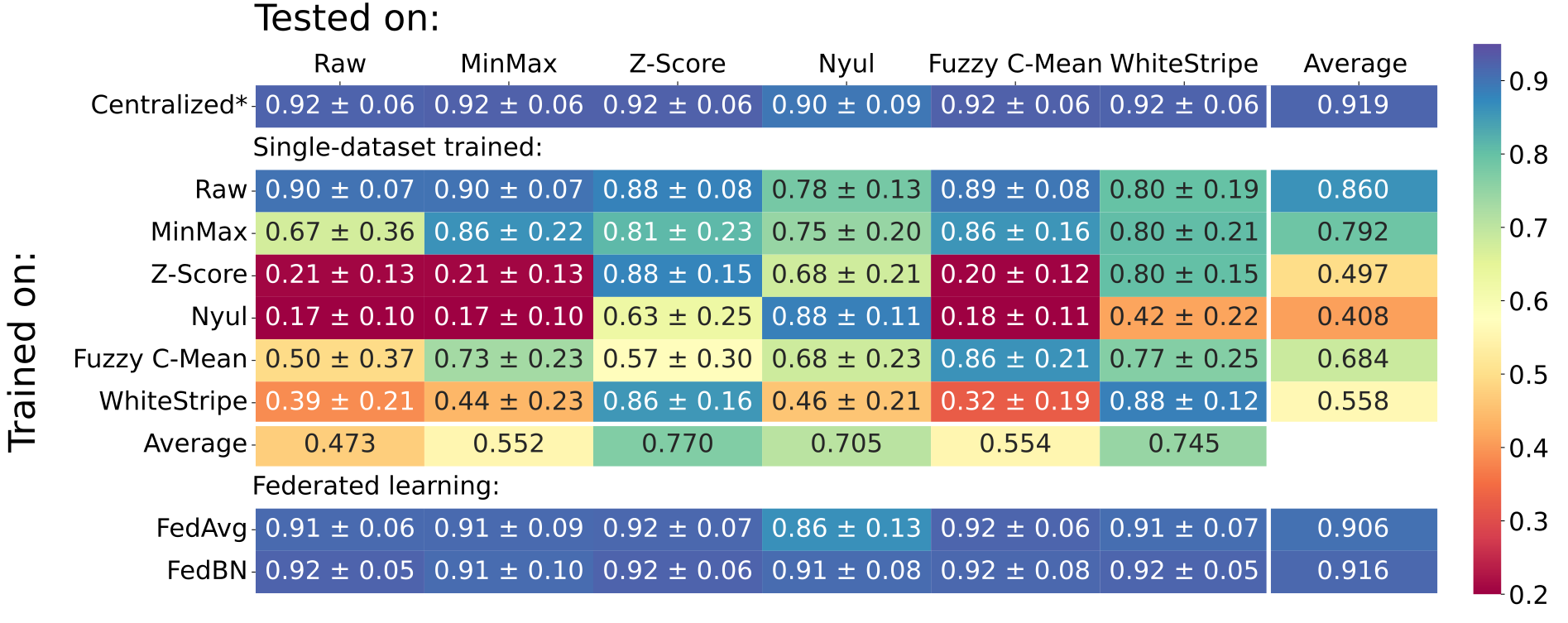}
\caption{3D Dice scores for all the models (rows) and all test sets (columns) with corresponding standard deviations. The star '*' next to the \textit{Centralized} model indicates that this model violates data privacy. The \textit{Average} row presents the average over the $GDS$ for single-dataset trained models, and the \textit{Average} column shows the average Dice among all the test sets for the given model.
} 
\label{fig:dice-table}
\end{figure}

The table in Figure \ref{fig:dice-table} gives a broad overview of the quality of segmentation models for all the variously normalized test sets. There is a clear diagonal for single-trained (ST) models, where high scores are obtained for models trained on identically normalized data. The highest average $GDS$ among the ST models was achieved for the model trained on the raw data. For some datasets, it performed even better than the model tested with the same normalization with which it was trained\footnote{This was possible thanks to the normalization layers. Without any normalization layers, there was no convergence during training on the raw dataset. However, training the same model with Z-score data was unstable, but in the end converged to similar loss values as for the model with Group Normalization.}.
However, this model convergence was unstable across different training runs, contrary to e.g. \textit{MinMax model}\footnote{For simplicity, "\textit{normalization\_name} model" is a short version of the phrase "model trained on data normalized with \textit{normalization\_name}" e.g. here it means model trained on data normalized with MinMax. Same for the datasets.}. 
The Nyul model performed the worst (probably because Nyul is the most aggressive normalization, see Figure \ref{fig_norm_histograms}). Notably, the results might be slightly biased due to different training samples.

\begin{figure}[p]
\includegraphics[width=\textwidth]{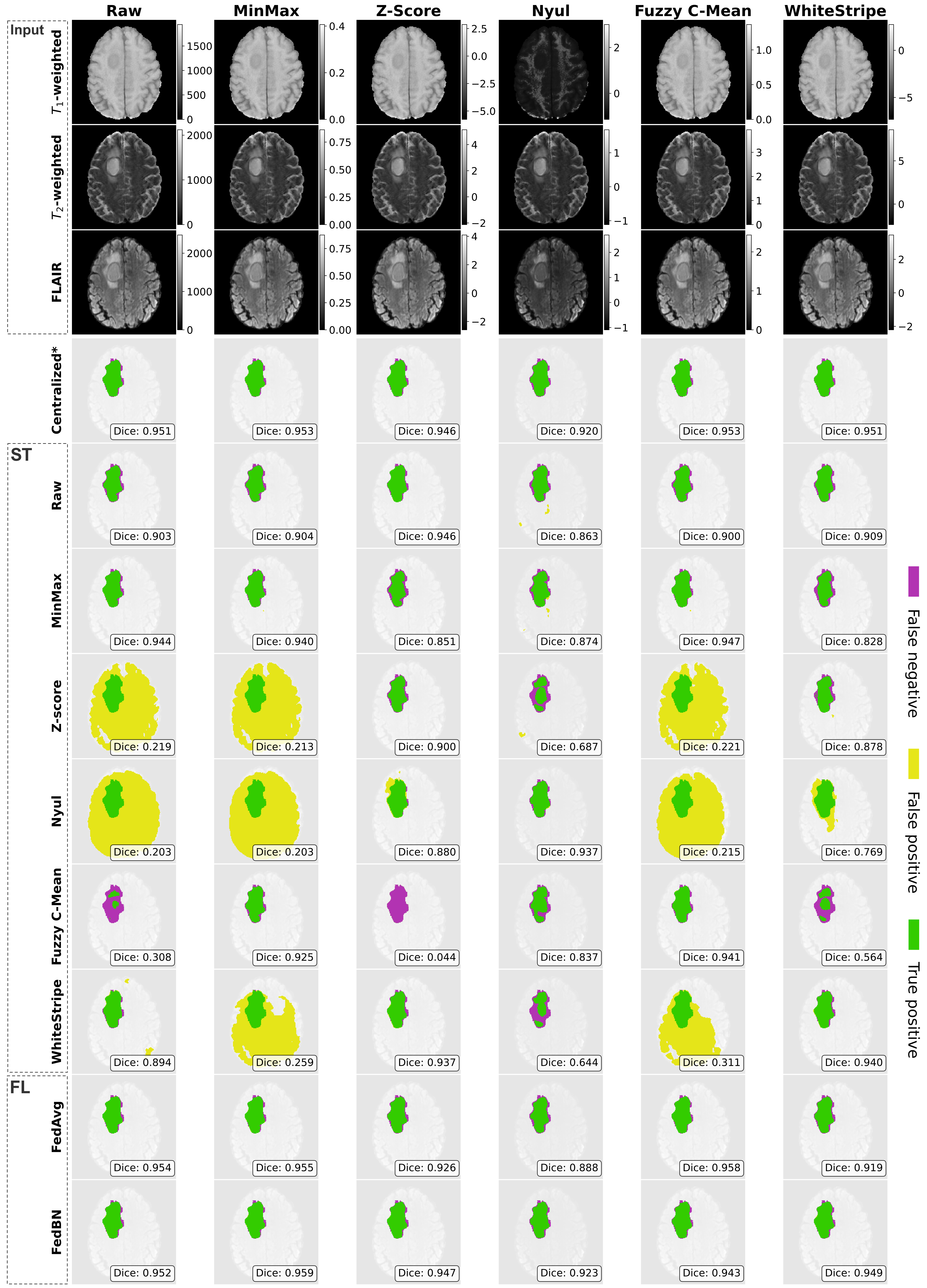}
\caption{Presentation of the predictions for each of the trained models on the 0229
patient. It presents the normalized inputs and the models’ predictions with distinction
for true positive, false negative, and false positive.
} 
\label{fig:preds-visualization}
\end{figure}

A similar pattern emerges in Figure \ref{fig:preds-visualization}. The predictions on the ST models diagonal are satisfactory (over 90\% of $GDS$ with a little of false negatives). The worst predictions overlap with the low score values in Fig. \ref{fig:dice-table}, such as Z-score, Nyul, and WhiteStripe models. They were usually oversegmenting, whereas the Fuzzy-C-Mean model suffered from the opposite, many false negatives (undersegmentation). Raw and MinMax performed very well, slightly worse than the best (centralized and FL) models.

Across all models, the centralized one achieves the highest score, although violating data privacy. The models trained with FL and the centralized one obtained comparable results. FedBN yielded similar $GSC$ to the centralized model within the margin of statistical error. The main difference between the FL methods was on the Nyul dataset, visible in both Figures \ref{fig:dice-table} and \ref{fig:preds-visualization}.  

Among the test sets, the Z-score normalized dataset turned out to be the most compatible with the other models (highest average score), but Fuzzy C-Mean model did not detect any tumor in the visualized slices. Also notable is the \textit{WhiteStripe dataset} with only a low $GDS$ for the worst model (Nyul). Furthermore, the raw datasets appeared to be the most challenging for all ST models. For FedAvg and the centralized models, the most problematic was the Nyul test set.

\section{Conclusions}

The results indicate some guidelines on training and running brain tumor segmentation models that can be potentially applied to other domains of deep learning and medical imaging. We could conclude that to train a well-generalizing model, which will be utilized for variously normalized data, the complicated normalization methods should be avoided and the diversity of the training set should be maintained. In the case that there are normalization layers in the neural network, the most optimal solution would be to leave the data raw. Contrary to what has been stated several times, neural networks do not have to be trained using only equally distributed data with feature values around zero (in the case of incorporating layer normalization)
\cite{ioffe2015batchnorm}.
% \cite{http://www.mysmu.edu/faculty/jingjiang/papers/da_survey.pdf}.

% Whereas, for the Z-Score trained model results were comparable to the presented ones.}.
Using an already trained model, the most promising approach is to replicate all preprocessing steps identically to the ones applied to the training data (the diagonal). Otherwise, scenarios like the one with the Nyul sample, where even well-trained models did not handle the input properly, may occur. However, if there is no information about the steps, the go-to normalization method should be Z-score (or WhiteStripe, which is a special type of Z-score). 

The federated learning models showed robustness for training with varying normalization methods between clients, achieving results comparable to the centralized model. 

However, the FL methods tested were the basic ones, and there is a possibility that a more sophisticated one would even outperform the centralized model. Nevertheless, the obtained results are satisfactory, and for better comparison, it might be needed to increase the complexity of the task or datasets. For example, the segmentation model could have some special context, such as 3D or 2.5D U-Net \cite{angermann20192.5unet,huang20203dunet}. Furthermore, leaving all four grades of glioma and performing multi-class segmentation would be another potential challenge for the model, exploiting the presented methods' proficiency.

\vfill

\section*{Acknowledgments}
The numerical experiment was possible through computing allocation on the Ares and Athena systems at ACC Cyfronet AGH under the grants PLG/2023/016117 and PLG/2024/016945. 

This project has received funding from the European Union's Horizon 2020 research and innovation programme under grant agreement No 857533 and from the International Research Agendas Programme of the Foundation for Polish Science No MAB PLUS/2019/13. 

The publication was created within the project of the Minister of Science and Higher Education "Support for the activity of Centers of Excellence established in Poland under Horizon 2020" on the basis of the contract number MEiN/2023/DIR/3796. 
% ---- Bibliography ----

% BibTeX users should specify bibliography style 'splncs04'.
% References will then be sorted and formatted in the correct style.

\bibliographystyle{splncs04}
% \bibliography{mybibliography}
\bibliography{references}

% \begin{thebibliography}{8}
% \bibitem{ref_article1}
% Author, F.: Article title. Journal \textbf{2}(5), 99--110 (2016)

% \bibitem{ref_lncs1}
% Author, F., Author, S.: Title of a proceedings paper. In: Editor,
% F., Editor, S. (eds.) CONFERENCE 2016, LNCS, vol. 9999, pp. 1--13.
% Springer, Heidelberg (2016). \doi{10.10007/1234567890}

% \bibitem{ref_book1}
% Author, F., Author, S., Author, T.: Book title. 2nd edn. Publisher,
% Location (1999)

% \bibitem{ref_proc1}
% Author, A.-B.: Contribution title. In: 9th International Proceedings
% on Proceedings, pp. 1--2. Publisher, Location (2010)

% \bibitem{ref_url1}
% LNCS Homepage, \url{http://www.springer.com/lncs}, last accessed 2023/10/25
% \end{thebibliography}
\end{document}